\documentclass[10pt,twocolumn,letterpaper]{article}

\usepackage{iccv}
\usepackage{times}
\usepackage{epsfig}
\usepackage{graphicx}
\usepackage{amsmath}
\usepackage{amssymb}
\usepackage{booktabs}
\usepackage{multirow}
\usepackage{bbm}
\usepackage{color}
\usepackage{amsmath}
\usepackage{mathtools}
\usepackage{subcaption}

\usepackage{booktabs}
\usepackage{algorithm}
\usepackage{algpseudocode}
\usepackage{multirow}
\usepackage{amssymb}
\usepackage{array}
\usepackage{graphicx}
\usepackage{makecell}
\usepackage{psfrag}
\UseRawInputEncoding

\usepackage[breaklinks=true,bookmarks=false]{hyperref}

\iccvfinalcopy 

\def\iccvPaperID{****} 

\ificcvfinal\fi

\begin{document}

\title{Unsupervised Person Re-identification via Simultaneous Clustering and Consistency Learning}

\author{
Junhui Yin, Jiayan Qiu, Siqing Zhang, Jiyang Xie, Zhanyu Ma$^{*}$, and Jun Guo
}

\maketitle
\ificcvfinal\thispagestyle{empty}\fi

\begin{abstract}
  Unsupervised person re-identification (re-ID) has become an important topic due to its potential
  to resolve the scalability problem of supervised re-ID models.
        However, existing methods simply utilize pseudo labels from clustering for supervision
        and thus have not yet fully explored the semantic information in data itself, which limits representation
capabilities of learned models.
   To address this problem, we design a pretext task for unsupervised re-ID by learning visual consistency from still images and temporal consistency during training process, such that the clustering network can separate the images into semantic clusters automatically. Specifically,
   the pretext task learns semantically meaningful representations by maximizing the agreement between two encoded views of the same image via a consistency loss in latent space. Meanwhile, we optimize the model by grouping the two encoded views into same cluster, thus enhancing the visual consistency between views. Experiments on Market-1501, DukeMTMC-reID and MSMT17 datasets demonstrate that our proposed approach outperforms the state-of-the-art methods by large margins.
\end{abstract}

\section{Introduction}

Convolutional Neural Networks
(CNNs) have undergone unprecedented success in visual
representation learning recently.
To realize rich representation, CNNs are typically trained using large-scale datasets (\emph{e.g.}, ImageNet~\cite{Deng2009ImageNet}) with extensive human annotations.
Despite their success, the large-scale data annotation is costly and laborious, especially for
complex data (\emph{e.g.}, videos) and concepts (\emph{e.g.}, image retrieval~\cite{2016Person,li2018harmonious,2019Unsupervised}).
In contrast to the methods \cite{krizhevsky2017imagenet,he2016deep} that rely on
heavy supervision, self-supervised learning is another kind of technique, which can easily obtain the supervision signal from the data itself and employ it to facilitate representation learning without
requiring extensive human annotation supervision.
\begin{figure}
\small
\begin{center}
\includegraphics[width=1.0\columnwidth]{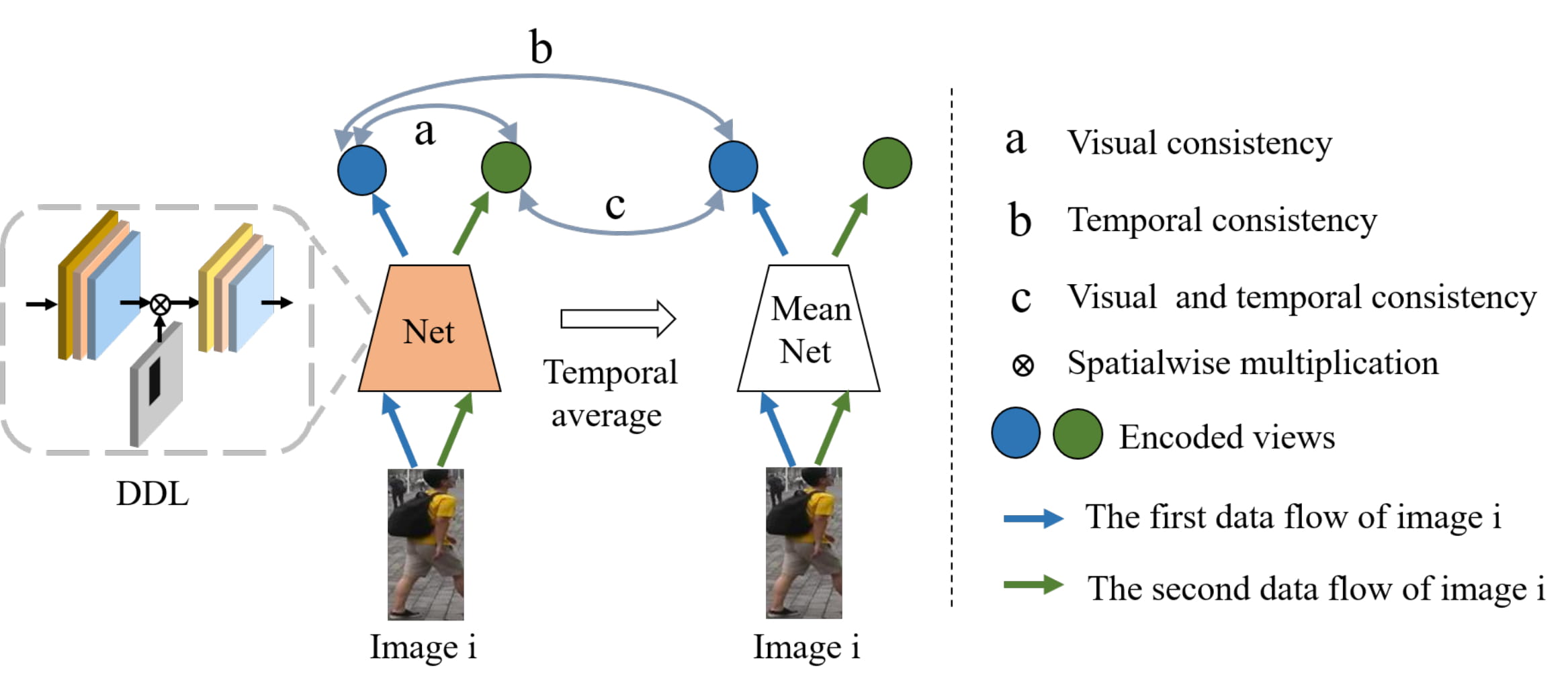}
\end{center}
   \caption{\small a: Visual consistency from two different encoded views of the same image based on network with Dynamic dropblock layer (DDL).
               b: Temporal consistency from the same encoded view at different stages (\emph{i.e.}, iterations).
               c: Visual and temporal consistency from different views at different stages.}
\label{fig:introduction1}
\end{figure}

A fertile source of annotation-free supervision is video since the visual world is continuous and does not change abruptly. Based on  this, \cite{wang2019learning} uses inherent consistency between observations adjacent in time as free supervisory information for training model.
Learning consistency meets the development of the human visual system as infants can
track and follow slow-moving objects before mapping objects to semantic meanings.
In this work, we focus on constructing consistency in still person images for unsupervised
representation learning under the framework of person re-identification (re-ID).


The requirement of re-ID is to match a person of interest with other images of this
person from non-overlapping cameras. Existing unsupervised re-ID approaches~\cite{fan2018unsupervised,lin2019bottom}
first extract embedding features of unlabeled dataset from
network model and then apply unsupervised clustering to divide
images into different clusters for training model.
These methods generate pseudo labels as ground-truth labels by simply using clustering algorithm and thus have not yet fully explored the semantic information existing in data itself, which limits representation capabilities of learned models. In addition,
the current clustering-based methods only focus on target task of re-ID and easily latch onto local information and low-level features
(e.g., color, contrast, texture, etc.), which is unwanted for the objective of semantic clustering \cite{van2014accelerating}.

To address the above problems, we
design a pretext task for unsupervised re-ID by learning consistency from still images, such that the clustering network can separate the images into semantic clusters rather than focusing on low-level features.
Similar to previous unsupervised representation learning \cite{wang2019learning,wang2020cycas}, the aim of \textbf{Cons}istency \textbf{Learn}ing (ConsLearn) is to learn feature representations that support reasoning
at visual and temporal consistency.
To construct the consistent information from single image and turn them into a learning signal for optimizing network,
we fisrt propose Dynamic Dropblock Layer (DDL), which randomly drops the region of
input maps, namely the semantic body parts, to obtain differently semantical representations based on the attentive feature
learning of remaining regions. Then, we feed the same person image into the model twice and apply DDL to convolutional feature maps of the model
to produce two encoded views (\emph{i.e.}, feature representations) of the image. Finally, the consistency across views is ensured by maximizing agreement between two encoded features
via a consistency loss in latent space.
The motivation behind ConsLearn
lies in that there is inherent visual consistency between the different views of the same images in space and the predicted target region of training image can still backtrace
to the initial information by using the information of the other view regardless of losing
the target on a view.

The above pretext task integrates disparate visual percepts into
persistent entities and enhances visual consistency in feature space.
To further enhance the temporal consistency across different training stages (\emph{i.e.}, iterations),
we improve
on existing methods by
adopting the
temporal average model of network from the previous iterations to generate the invariance
feature for a view of image, which
enables the consistency with the other view obtained by the current encoder model, as shown in Figure \ref{fig:introduction1}.


Simultaneous clustering and consistency learning is one of the most promising
approaches for unsupervised learning of deep neural networks.
As a pretext task, ConsLearn can obtain the free supervision
for consistency from the unlabeled data and learn feature representations
that supports identifying consistency across views.
As these features improve, the ability to predict the missing patch of feature maps are enhanced,
inching the model toward consistency.
As a target task, clustering process trains a deep neural network using pseudo-labels and encourages network model to
focus on the current task of re-ID. The self-supervised task from representation learning requires semantic understanding
and thus induces clustering to obtain semantically meaningful features rather than depending on low-level features, which is
present in existing clustering-based approaches~\cite{fan2018unsupervised,lin2019bottom,fu2019self}.

The main contributions of this work can be
summarized as follows:

$\bullet$ A novel \textbf{Cons}istency \textbf{Learn}ing (ConsLearn) approach is proposed to construct data association across encoded views from unlabeled pedestrian images
and learn representations for visual reasoning in a self-supervised manner.

$\bullet$ ConsLearn is implemented with clustering, which not only perceives the current pretext task, but also learns the feature information related to the target task.
In this way, our method can automatically separate images into semantically
meaningful clusters.



$\bullet$ The proposed method is applied on unsupervised person re-ID and further extended to unsupervised domain adaptation (UDA) re-ID,
and improves the state-of-the-arts with significant
margins.

\section{Related Work}

\textbf{Unsupervised person re-identification}. Unsupervised person re-identification (re-ID)
is very challenging to learn discriminative representation due to the absence of target labels as learning guidance.
Several methods \cite{fan2018unsupervised,lin2019bottom,zeng2020hierarchical} resort to separate
unlabeled data into different clusters by clustering algorithms (\emph{e.g.}, DBSCAN \cite{ester1996density}) and iteratively trains network models based on the generated pseudo labels. However, these methods highly rely on pseudo labels and may be affected due to the hard quantization error caused by the clustering error. To address this issue, some existing methods
use some weak supervised signal for training re-ID model. For instance, the tracklet-based method \cite{li2018unsupervised} uses person tracklet as supervised information to jointly learning within-camera tracklet correlation and cross-camera tracklet association, without any additional annotations. Despite their success, these methods are not truly unsupervised learning in essence
and work hard when there are not labeled data and weak supervised information (\emph{e.g.}, tracklet).
Another line of this work is to
employ deep re-ID model learned from labeled source data as an initialized feature extractor and adapt the model with both source domain and target domain to reduce the domain discrepancy between different datasets on image-level \cite{deng2018image,wei2018person,bak2018domain} and attribute feature-level \cite{liu2019adaptive,chen2019instance}.
\cite{ge2020mutual} and~\cite{zhai2020multiple} use multiple network models and their corresponding mean nets to generate soft pseudo labels for supervising other models. The main difference from them is that we construct temporal consistency during training
process by only using one current network and its past
(mean) network with DDL. In addition, many other contrastive learning methods~\cite{he2020momentum,grill2020bootstrap} also using the momentum-based averaging model.

\begin{figure*}
\begin{center}
\includegraphics[width=1.9\columnwidth]{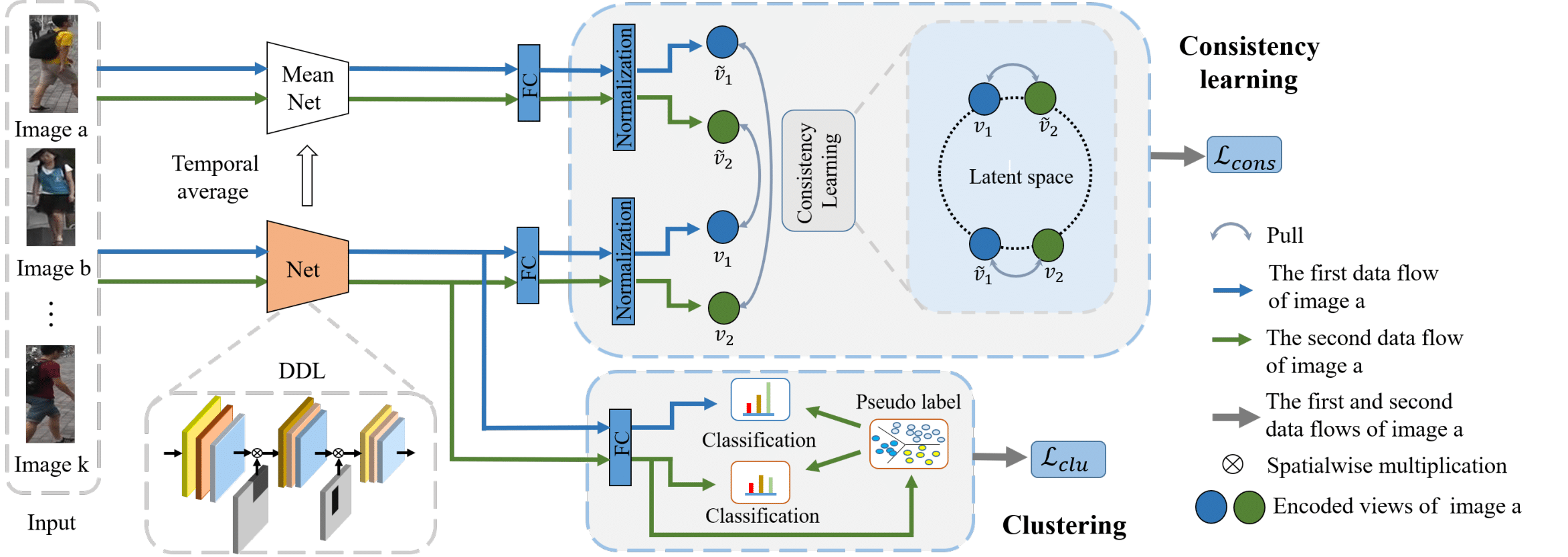}
\end{center}
   \caption{\small Illustration of our proposed framework. During training,
  unlabeled data are fed-forward into the
deep re-ID network with DDL and its temporally averaging model to obtain differently encoded views (\emph{i.e.}, feature representations).
Subsequently, two components are designed to optimize the network with unlabeled data, respectively.
The first component is a clustering module that provides the same cluster assignments for two encoded features to calculates the cross-entropy loss. The second component is ConsLearn module that learns vision and temporal consistency from still images via a consistency loss.
}
\label{fig:overview}
\end{figure*}


\textbf{Self-supervised representation learning}. As a form of unsupervised learning,
self-supervised representation learning is
a kind of techniques that learn representations by solving
a pretext task, where the supervision signal can be
obtained freely from the large amount of unlabeled
data available online.  
Contrastive learning~\cite{hadsell2006dimensionality} is
the core idea in this line of investigation.
It
has recently shown great progress~\cite{wu2018unsupervised,tian2019contrastive,henaff2020data,bachman2019learning,ye2019unsupervised}. As a milestone,
MoCo~\cite{he2020momentum} maintains a queue of negative
samples and train the model with a memory bank to improve consistency of the queue.
After that,
considerable attention
has been focused in this direction~\cite{chen2020simple,tian2020makes,cao2020parametric,grill2020bootstrap,caron2020unsupervised}
and spread
over many other tasks~\cite{xie2020propagate,han2020self}. Inspired by MoCo series~\cite{he2020momentum,chen2020simple},
we use the mean net to construct temporal consistency
during training process.
Differen form these works, our method is
The main difference from MoCo series~\cite{he2020momentum,chen2020simple} is our method can be thought of as ¡°MoCo
without negatives¡± and rarely depend on large batch, while MoCo highly relies on them.


Compared to still images, videos should be a fertile source for self-supervised
learning as they naturally possess the temporal resolution
and dynamics.
A suitable supervision for free is inherent visual
correspondence between adjacent frames in video. 
\cite{wang2019learning} utilizes cycle consistency as free supervision to learn fine-grained correspondences between pixels.
\cite{wang2020cycas} propose a data association method, which concentrates more on learning high-level semantic correspondences and is adaptive to the re-ID problem. However, this technique is required to the prior knowledge of camera view labels, which sometimes are unavailable.
The above methods usually focus on learning consistency between adjacent frames of video, while ours concentrate more on
how to find the consistency in still person images and utilize it to provide a self-supervised solution for person re-ID task, without any additional information (\emph{e.g.}, cameral view labels).

\section{Approach}

In this section, we first propose
the consistency learning (ConsLearn) as a pretext task
for unsupervised person re-ID. We then introduce the discriminative
clustering method and incorporate it with ConsLearn in a joint optimization framework. The overview of the training procedure is illustrated in Figure \ref{fig:overview}.

\subsection{Consistency Learning}

Clustering tends to focus on low-level features, like color, which
is unessential for the objective of semantic clustering. To explore richer
visual patterns and obtain semantically meaningful features, we propose ConsLearn and integrate it with the clustering network.
The following sections present the cornerstones of our approach, which can
encourage clustering network to separate images into semantically meaningful
clusters without human annotations.

\textbf{Dynamic dropblock layer.}
In this subsection, we present the details of dynamic dropblock layer (DDL).
 DDL can be applied
on any feature layer of network model, and induce model to capture the different semantic information of the object during
training. It is worth noting that DDL is deactivated during the testing phase.

The input of DDL is a convolutional feature
map $F\in\mathbb{R}^{H\times W\times C}$ where $C$ represents the number of channels,
$H$ and $W$ are height and width of feature
map, respectively.
We first set two main hyperparameters $\alpha$ and $\beta$. They control
the height and width ratio of the erased region on map $F$, respectively.
The size of region to be dropped increases as $\alpha$ and $\beta$ increases and is adaptive to the size of feature map when the hyperparameters are fixed. Therefore, the dropblock layer is dynamic.
The height and width of the erased region vary
from task to task. But in our work, the removed region should contain a semantic part of feature map.
Specifically,
the drop mask $M_{drop}\in\mathbb{R}^{H\times W}$ is generated
by setting the pixel to 0 if it is inside the generated erased region,
and 1 otherwise.
The DDL randomly remove the
part of feature
map $F$ by multiplying it to the drop mask.
In this way, we can produce different views for input feature map, namely different semantic body parts.
We visualize the example
of each component in the following experiment. The regularization technique of Cutout~\cite{devries2017improved}
also studies the application of random erasing. A difference from Cutout is
the erased region size of our DDL changes dynamically
with the size of feature map, while the removed region
size of Cutout is fixed and squared. The other
difference is that our DDL can be applied on any feature layer
of network instead of only input image as in Cutout. In fact, our DDL can be
considered as an extension of random erasing for feature maps and also an extension of Cutout~\cite{devries2017improved} or Dropout~\cite{srivastava2014dropout}.

\textbf{Learning consistency}.
Consider as input a person images $x$, the pixel input is mapped to a
feature space twice by an encoder with DDL, such that $h_{k}=\Phi_{k}(x|\theta)$, $k=1,2$.
Here, $k=1,2$ mean the two data flows of the same image,
which are fed into network in order.
Since the erased area of DDL is given randomly at each layer,
the firstly entered network $\Phi_{1}(\cdot|\theta)$ is different from the second network $\Phi_{2}(\cdot|\theta)$.
Thus, $k$ is also used to distinguish them. This is also why we use
two data flows for the same image.
In this way,
the network models $\Phi_{k}(\cdot|\theta)$ output two different encoded views $h_{k}$, $k=1,2$, leading to
two differently semantic features from the same image.
Our aim is to
reduce the ability for clustering depending on low-level features and induce the model
to learn the inherent visual consistency across different views. Therefore, we want to maximize the agreement between $h_{1}$ and $h_{2}$.
A linear layer $g$: $\mathbb{R}^{D}\rightarrow\mathbb{R}^{\Omega}$ follows the feature representations to map them to a latent space,
which is defined as
\begin{equation}\label{tik}
\small
u_{k}=g(h_{k}), k=1,2.
\end{equation}
In this space, our model can better learn a semantically meaningful representation that describes what is in common
between $u_{1}$ and $u_{2}$ while discarding instance-specific details.

The encoder model at current iteration can output the encoded views for all the person images, but this
rapidly changing encoder also reduces the
feature representations' consistency across different training stages and eventually results in training
error amplification. To resolve this issue, we present the momentum-based moving average
model
as a way of preserving more original knowledge for unsupervised learning with a consistency loss.
Formally, let the temporal average model be $\Phi_{k}^{*}(\cdot|\theta^{*})$. Then, its model parameter $\theta^{*(t)}$ at current iteration $t$
can be updated by
\begin{equation}\label{momentum}
\small
\theta^{*(t)}=\zeta\theta^{*(t-1)}+(1-\zeta)\theta^{(t)},
\end{equation}
where $\zeta\in[0,1)$ is a momentum coefficient and the initial temporal average parameters $\theta^{*(0)}$ is $\theta$.

Two momentum-based encoded views of the same images are generated by the temporal average model $\phi$ and the linear layer $g^{*}$.
That is,
$\tilde{u}_{k}=g_{k}(\Phi_{k}^{*}(x_{k})), k=1,2$.
Considering the errors caused by self-variation and large differences in values, the feature distributions $u_{k}$ and $\tilde{u}_{k}$ are scaled up by a softmax operation,
\begin{equation}\label{tik}
\small
v_{kl}=\delta(u_{k})=\frac{e^{u_{kl}}}{\sum_{j=1}^{\Omega}e^{u_{kj}}}, k=1,2,
\end{equation}
\begin{equation}\label{tik}
\small
\tilde{v}_{kl}=\delta(\tilde{u}_{k})=\frac{e^{\tilde{u}_{kl}}}{\sum_{j=1}^{\Omega}e^{\tilde{u}_{kj}}}, k=1,2,
\end{equation}
where $v_{kl}(\tilde{v}_{kl})$ is the $l$-th element of $v_{k}(\tilde{v}_{k})$.
Now, the problem is how to turn the consistency information into a
learning signal for optimizing network models.

It is well known that the Kullback-Leibler (KL) divergence $\displaystyle KL(\cdot)$ can be thought of as a measurement of how far the distribution $Q$ is from the distribution $P$.
For discrete probability distributions $P$ and $Q$ defined on the same probability space $\mathcal{X}$,
$P$ typically represents the ``true" distribution of data, while $Q$ typically represents a approximation of $P$. In order to find a distribution $Q$ that is closest to $P$, we can minimize $KL$ divergence, which can be rewritten by
\begin{equation}\label{tik}
\small
\begin{aligned}
KL(P\|Q) &=-\sum_{x\in\mathcal{X}}P(x)\log(Q(x))+\sum_{x\in\mathcal{X}}P(x)\log(P(x)) \\
&=H(P,Q)-H(P),
\end{aligned}
\end{equation}
where $H(P,Q)$ is the cross entropy of $P$ and $Q$, and $H(P)$ is the entropy of $P$. Here, $KL(P\|Q)$ is proportional to $H(P,Q)$ if $H(P)$ is a constant.

The embedding features $v_{k}$ and $\tilde{v}_{k}$, $k=1,2,$ can
be interpreted as the distribution of a discrete random variable over training samples $x$. Furthermore, $\tilde{v}_{i}$ can be viewed as a constant or ``true" distribution since it represents invariant feature distribution obtained by the temporally moving average model. Our aim is that $v_{1}$ should be close to $\tilde{v}_{2}$ and $v_{2}$ should be close to $\tilde{v}_{1}$ as much as possible, which supports visual reasoning between different semantic features. Therefore, $H(P,Q)$ is adopted to turn the consistency information into a learning signal for optimizing our models.
The consistency of different encoded views is measured using
the following loss, as
\begin{equation}\label{cons}
\small
\begin{aligned}
\mathcal{L}_{co} &= H(\tilde{v}_{1},v_{2})+H(\tilde{v}_{2},v_{1})  \\
&=-\sum\tilde{v}_{1}\log(v_{2})-\sum\tilde{v}_{2}\log(v_{1}).
\end{aligned}
\end{equation}

\subsection{Unsupervised Person Re-identification by Clustering}
Unlike pervious clustering techniques~\cite{fan2018unsupervised,lin2019bottom,fu2019self},
our clustering method first generates pseudo labels according to features from a view of the same images and utilizes the same labels for all the encoded views since they share large visual similarity. Its aim is to further enhance the visual consistency between different views based on unsupervised clustering. Specifically,
given encoded features $\{h_{1}^{i}\}_{i=1}^{N}$ of images $\{x^{i}\}_{i=1}^{N}$ from a view, clustering algorithm (\emph{i.e.}, DBSCAN~\cite{ester1996density}) divides these learned features into different clusters as pseudo label $\{\hat{y}\}_{i=1}^{N}$, which represents the image's membership to one of $M$ possible predefined categories.
Meanwhile, the representations are also fed into
a linear classification layer, converting
the feature vector $h_{1}^{i}$ into a vector of class scores $z_{1}^{i}$.
Finally, the softmax operator is performed to map the class scores to class probabilities $p(\hat{y}_{i}|x_{i})$, which is the predicted probability of image $x_{i}$ belonging
to the identity $\hat{y}_{i}$.
As in~\cite{zhong2019invariance},
the clustering procedure for two encoded views is achieved by minimizing the cross entropy loss,
\begin{equation}\label{clustering}
\small
\mathcal{L}_{ce}= -\frac{1}{N}\sum_{i=1}^{N}\log p(\hat{y}_{i}|h_{1}^{i})-\frac{1}{N}\sum_{i=1}^{N}\log p(\hat{y}_{i}|h_{2}^{i}).
\end{equation}

Furthermore, we introduce a softmax-triplet loss to enforce a positive pair to be closer and a negative
pair apart from each other in the representation space.
As in \cite{hermans2017defense}, let $h_1^{(in)}$ and $h_1^{(ip)}$ denote the similarity of the hardest negative pair $\{h_1^{i}, h_1^{n}\}$ and the hardest positive pair
$\{h_1^{i}, h_1^{p}\}$ for encoded views $\{h_{1}^{i}\}_{i=1}^{N}$, with an anchor $h_1^{i}$. The similarity can be computed as
$h_1^{(in)}=-\|h_1^{i}-h_1^{n}\|$ and
$h_1^{(ip)}=-\|h_1^{i}-h_1^{p}\|$,
where $\|\cdot\|$ represents Euclidean distance.
In this sense, minimizing the similarity of the hardest negative pair means maximizing the distance between $h_1^{(i)}$ and $h_1^{(n)}$ and vice versa. Similarity, encoded views $\{h_{2}^{i}\}_{i=1}^{N}$ have similar results.
Thus, the softmax-triplet loss for two encoded views can be performed by
\begin{equation}\label{triplet}
\small
\begin{aligned}
\mathcal{L}_{st}= & -\frac{1}{N}\sum_{i=1}^{N}\log \frac{\exp(h_1^{(in)})}{\exp(h_1^{(in)})+\exp(h_1^{(ip)})} \\
& -\frac{1}{N}\sum_{i=1}^{N}\log \frac{\exp(h_2^{(in)})}{\exp(h_2^{(in)})+\exp(h_2^{(ip)})}.
\end{aligned}
\end{equation}

Overall, unsupervised deep clustering alternates between clustering features as pseudo label to update the parameters of the model
using Eq.~\eqref{clustering} and learning features by collecting and distinguishing informative
pairs accurately using Eq.~\eqref{triplet}.
The proposed procedure can be achieved by
\begin{equation}\label{clustering+st}
\small
\begin{aligned}
\mathcal{L}_{clu}=\lambda\mathcal{L}_{ce}+\xi\mathcal{L}_{st}
\end{aligned}
\end{equation}

\textbf{The overall loss for network}. ConsLearn is implemented with clustering, which not
only perceives the current pretext task, but also learns the
feature information related to the target task.
By adding the clustering loss (Eq.~\eqref{clustering+st}) into the consistency loss (Eq.~\eqref{cons}), the overall
loss $\mathcal{L}$ of our final proposed method can be expressed as
\begin{equation}\label{objective}
\small
\mathcal{L}=\lambda\mathcal{L}_{ce}+\xi\mathcal{L}_{st}+\eta\mathcal{L}_{co},
\end{equation}
where $\lambda$, $\xi$ and $\eta$ are weighting parameters.

\section{Experiments and Disscussions}

\subsection{Experimental Settings}
\textbf{Datasets and evaluation protocol}. Our proposed model is evaluated on three public person re-identification (re-ID) benchmarks:
Market-1501 \cite{zheng2015scalable}, DukeMTMC-reID \cite{ristani2016performance}, and MSMT17 \cite{wei2018person}. To evaluate the performance of our method, two evaluation metrics are chosen
as the mean Average Precision (mAP) and the cumulative matching characteristic (CMC)
top-1, top-5, top-10 accuracies ~\cite{zheng2015scalable}.

\textbf{Implementation details}.
All person images are resized to 256$\times$128 before fed into the networks. Then, we adopt random cropping and flipping as the data augmentation. After that, we feed these images into ResNet-50 \cite{he2016deep} with the last classification layer removed, initialized by the ImageNet \cite{Deng2009ImageNet}.
We use a mini-batch size of 256 in 4
GPUs. The network models are trained for 50 epochs, using ADAM optimizer with learning
rate of 0.00035, weight decay of 0.0005. DBSCAN \cite{ester1996density} is used for clustering
in each epoch.
For the momentum coefficient, we set $\zeta$ as 0.999.
We empirically set the hyperparameters $\lambda, \xi$ and $\eta$ for loss balance as 0.2, 0.35 and 0.1 for unsupervised re-ID on benchmarks including Market-1501 and DukeMTMC-reID.
For other datasets, these parameters may be slightly different.

\begin{table}[t]
\footnotesize
  \centering
    \begin{tabular}{ccc|cc|cc}
    \toprule
    \multicolumn{1}{c|}{\multirow{2}[4]{*}{Methods}} & \multicolumn{2}{c|}{Market} & \multicolumn{2}{c|}{Duke}  & \multicolumn{2}{c}{MSMT17} \\
\cmidrule{2-7}    \multicolumn{1}{c|}{} & \multicolumn{1}{c|}{mAP} & \multicolumn{1}{c|}{top-1}  & \multicolumn{1}{c|}{mAP} & \multicolumn{1}{c|}{top-1}
& \multicolumn{1}{c|}{mAP} & \multicolumn{1}{c|}{top-1} \\
    \midrule
    \midrule

    \multicolumn{1}{l|}{OIM ~\cite{xiao2017joint}}   & 14.0  & 38.0     & 11.3    & 24.5    & - & -  \\
    \multicolumn{1}{l|}{BUC ~\cite{lin2019bottom}}   & 38.3  & 66.2      & 27.5    & 47.4  & - & -  \\
    \multicolumn{1}{l|}{SSL ~\cite{lin2020unsupervised}}   & 37.8  & 71.7     & 28.6    & 52.5   & - & -  \\
    \multicolumn{1}{l|}{MAR ~\cite{yu2019unsupervised}}   & 40.0  & 67.7    & 48.0    & 67.1   & - & -  \\
    \multicolumn{1}{l|}{TAUDL ~\cite{fan2018unsupervised}} & 43.5  & 61.7     & 41.2    &  63.7    & 12.5 & 28.4  \\
    \multicolumn{1}{l|}{MMCL ~\cite{wang2020unsupervised}}  & 45.5  & 80.3    & 40.2    & 65.2   & 11.2 & 35.4  \\
    \multicolumn{1}{l|}{HCT ~\cite{zeng2020hierarchical}}  & 56.4  & 75.2     & 45.1    & 63.2   & - & -  \\
    \multicolumn{1}{l|}{CycAs ~\cite{wang2020cycas}}  & 64.8  & 84.8   & 60.1    & 77.9     & - & -   \\
    \multicolumn{1}{l|}{MMT$^{*}$ ~\cite{ge2020mutual}}  & 74.7  & 88.4    & 60.8    & 75.0    & 25.8 & \textbf{57.2}  \\   	
    \midrule
    \multicolumn{1}{c|}{Ours} & \textbf{77.7} & \textbf{90.4}  & \textbf{64.5} & \textbf{78.1}  & \textbf{27.7} & 55.5  \\
    \bottomrule
    \end{tabular}%
    \vspace{0.1cm}
  \caption{\small Comparisons with state-of-the-art methods on three benchmarks for unsupervised person re-ID task. ``*" means the results are reproduced by implementing the authors' code. Bold text refers the best performance.}
  \label{tab:duke market and msmt17}%
\end{table}%

\begin{table}[htbp]
\small
  \centering
  \scalebox{0.93}{
    \begin{tabular}{cccc|ccc}
    \toprule
    \multicolumn{1}{c|}{\multirow{2}[4]{*}{Methods}} & \multicolumn{3}{c|}{Duke $\rightarrow$ Market} & \multicolumn{3}{c}{Market $\rightarrow$ Duke }   \\
\cmidrule{2-7}
  \multicolumn{1}{c|}{} & \multicolumn{1}{c}{mAP} & \multicolumn{1}{c}{top-1} & \multicolumn{1}{c|}{top-5} & \multicolumn{1}{c}{mAP} & \multicolumn{1}{c}{top-1} & \multicolumn{1}{c}{top-5} \\
    \midrule
    \midrule
    \multicolumn{1}{l|}{PUL \cite{fan2018unsupervised}}  & 20.5  & 45.5  & 60.7    & 16.4  & 30.0    & 43.4   \\
    \multicolumn{1}{l|}{SPGAN \cite{deng2018image}}  & 22.8  & 51.5  & 70.1    & 22.3  & 41.1  & 56.6  \\
    \multicolumn{1}{l|}{HHL \cite{zhong2018generalizing}} & 31.4  & 62.2  & 78.8   & 27.2  & 46.9  & 61.0   \\
    \multicolumn{1}{l|}{UCDA \cite{qi2019novel}}  & 30.9  & 60.4  &   -       & 31.0    & 47.7  &   -    \\
    \multicolumn{1}{l|}{PDA-Net \cite{li2019cross} } & 47.6  & 75.2  & 86.3   & 45.1  & 63.2  & 77.0     \\
    \multicolumn{1}{l|}{CR-GAN \cite{chen2019instance}} & 54.0    & 77.7  & 89.7   & 48.6  & 68.9  & 80.2   \\
    \multicolumn{1}{l|}{PCB \cite{zhang2019self}}  & 54.6  & 78.4  &    -     & 54.3  & 72.4  &  -  \\
    \multicolumn{1}{l|}{SSG \cite{fu2019self}}  & 58.3  & 80.0    & 90.0   & 53.4  & 73.0    & 80.6  \\
    \multicolumn{1}{l|}{ECN++ \cite{zhong2020learning}}  & 63.8  & 84.1  & 92.8   & 54.4  & 74.0    & 83.7 \\
    \multicolumn{1}{l|}{MMCL \cite{wang2020unsupervised}}  & 60.4  & 84.4  & 92.8     & 51.4  & 72.4  &  82.9   \\
    \multicolumn{1}{l|}{SNR \cite{jin2020style}}  & 61.7  & 82.8  &    -     & 58.1  & 76.3  &  -    \\
    \multicolumn{1}{l|}{AD-Cluster \cite{zhai2020ad}} & 68.3  & 86.7  & 94.4   & 54.1  & 72.6  & 82.5  \\
    \multicolumn{1}{l|}{MMT \cite{ge2020mutual}}  & 71.2  & 87.7  & 94.9  & 65.1  & 78.0    & 88.8   \\
    \multicolumn{1}{l|}{DG-Net++ \cite{zou2020joint}}& 61.7  & 82.1  & 90.2   & 63.8 &  78.9 &  87.8  \\
    \multicolumn{1}{l|}{MEB-Net \cite{zhai2020multiple}}  & 76.0  & 89.9  & 96.0   & 66.1  & 79.6    & 88.3 \\
    \midrule
    \multicolumn{1}{c|}{Ours} & \textbf{82.2} & \textbf{92.7} & \textbf{96.9}  & \textbf{68.3} & \textbf{80.9} & \textbf{90.2} \\
    \bottomrule  			

    \end{tabular}}%
    \vspace{0.05cm}
      \caption{\small Comparisons with state-of-the-art methods on Dukemtmc-reID (Duke) and Market-1501 (Market) for domain adaptive tasks.}
  \label{tab:duke and market}%
\end{table}%

\subsection{Comparisons with State-of-the-arts}

\textbf{Unsupervised person re-identification}. To verify effectiveness of the proposed method, we compare our method with the state-of-the-art methods by training the re-ID
model without any labeled data. Table \ref{tab:duke market and msmt17} shows the experimental results on three benchmarks, including
Market-1501 (Market), DukeMTMC-reID (Duke), and MSMT17.
On Market-1501, under the same setting, our method achieves the best performance
among the compared methods with mAP=77.7\% and top-1=90.4\%, which exceeds the state-of-the-art method (MMT \cite{ge2020mutual}) by large margins (3.0\% for mAP and 2.0\% for top-1). 
On DukeMTMC-reID, we also achieve evident 3.7\% mAP improvement over the current best approach.
For a larger and more challenging dataset (MSMT17), the proposed method has also achieved comparable performance in terms of mAP and CMC.
With the obtained results, we validate that
our method is able to learn richer representations
of person images compared with previous methods.

\textbf{Unsupervised domain adaptation person re-identification}. To further demonstrate its effectiveness,
our method is tested on several domain adaptation tasks, which are the representative
target tasks to validate the effectiveness of unsupervised representation. These tasks contains Duke-to-Market and Market-to-Duke. We initialize the
backbone with the model pre-trained on labeled source domain, and fine-tune on unlabeled target domain.

Table \ref{tab:duke and market} summarizes the experimental results on two adaptation tasks.
On Market-to-Duke, compared to the state-of-the-art method, our method obtains 6.2\% and
2.8\% improvements on mAP and Top-1 accuracy, respectively. Similar results
can be observed in Market-to-Duke.
 The above impressive performances demonstrate the necessity and effectiveness of our proposed method for unsupervised domain adaptation re-ID. More importantly, current advanced techniques (MMT  ~\cite{ge2020mutual} and MEB-Net \cite{zhai2020multiple}) train multiple networks
to enhance the discrimination capability of re-ID model, while our method only use a network and thus is not
required for the additional learning parameters in training stage compared to MMT and MEB-Net.

\subsection{Ablation Studies}

In this section, we conduct extensive experiments on DukeMTMC-reID to analyze the effect of our method under different
setting.


\begin{table}[t]
  \centering
  \small
    \begin{tabular}{l|cccc}
    \toprule
    \multicolumn{1}{l|}{Applied stage}  & mAP & top-1 & top-5 & top-10  \\
    \midrule
    $stage\_\{0\}$ & 63.0  &  77.7  & 87.3  & 90.6   \\
    $stage\_\{1,3\}$  & 63.3 & 77.8  & 87.5  & 90.7   \\   		
    $stage\_\{0,1,2\}$ & \textbf{64.5}  & \textbf{78.1}  & \textbf{87.8}  & \textbf{90.9}   \\

    $stage\_\{0,1,2,3,4\}$  & 61.8 & 75.9  & 85.8  & 89.8   \\
    $stage\_\{2,3,4\}$ & 59.3  & 75.3  & 84.9  & 88.3   \\ 	
    $stage\_\{2,4\}$  & 58.4 & 74.7  & 84.7  & 87.7   \\
    $stage\_\{3,4\}$  & 57.7 & 74.3  & 84.2  & 87.2   \\
            \midrule
    \multicolumn{1}{l|}{N/A} & 54.2  & 72.0    & 81.3 & 85.1  \\
    \bottomrule
    \end{tabular}%
    \vspace{0.1cm}
      \caption{\small Effects in performance upon the choice of the feature maps to
employ DDL. $stage\_\{0,1,2\}$ refers that DDL is plugged into network before the intermediate stages: $stage$\_0, $stage$\_1, and $stage$\_2,
and others are similar. N/A indicates that
DDL outputs the raw input feature map instead of applying DDL.
}
  \label{tab:applied stage}%
\end{table}%

\begin{table}[t]
  \centering
     \small
    \begin{tabular}{p{1.8cm}|p{1.5cm}<{\centering}|p{0.67cm}<{\centering}p{0.67cm}<{\centering}p{0.67cm}<{\centering}}
    \toprule
    \multicolumn{1}{l|}{Applied stage}  &  $(\alpha,\beta)$ & mAP & top-1 & top-5  \\
    \midrule
    \multirow{4}{*}{$stage\_\{0,1,2\}$}   & (0.2,0.1)  & 56.0 & 73.0    & 82.9  \\
          & (0.3,0.2) & 60.6  & 75.9  & 86.0  \\
   & (0.4,0.3)  & \textbf{64.5} & \textbf{78.1}  & \textbf{87.8}  \\
          & (0.5,0.4) & 56.4 & 71.2  & 83.1  \\
    \midrule
\multirow{4}{*}{$stage$\_\{3,4\}}   & (0.2,0.1) & 55.4 & 72.4  & 82.4   \\
              & (0.3,0.2)  & 56.8 & 73.9  & 84.0   \\
          & (0.4,0.3)  & 57.7 & 74.3  & 84.2 \\
          & (0.5,0.4) & 55.8 & 73.0  & 83.3   \\
    \bottomrule
    \end{tabular}%
    \vspace{0.1cm}
    \caption{\small Accuracies according to the size $(\alpha,\beta)$ of the erased region.
     Upper: Accuracies with DDL deactivated.
     Middle: Accuracies when DDL applied for lower-level layers.
  lower: Accuracies when DDL applied for higher-level layers.}
\label{tab:11}%
\end{table}%

\textbf{Dynamic dropblock layer}.
First, we explore the influence of DDL at different stages by
integrating DDL into ResNet-50. During the training phase, we plug DDL into network before the intermediate stages:
stage 0, stage 1, stage 2, stage 3, and stage 4.
For instance, stage 0 means that DDL is plugged into network before input map is fed into network.
Table \ref{tab:applied stage} reports the experimental results. From
these results, it can be observed that the best accuracy can be achieved when the DDL is plugged in lower-level layers. However, the performance of DDL becomes worse under higher-level layers. The reason behind is that first-layer features are typically more general (\emph{e.g.}, unrelated to a particular task) while last-layer features exhibit greater levels of specificity (\emph{e.g.}, semantic and appearance diversity). This indicates that DDL tends to work if the features are general, meaning suitable to both pretext task and re-ID task.
In addition,
as the last line of Table~\ref{tab:applied stage}, the performance of our method (w/o DDL) is worse than that of our method, which demonstrates that our method benefit from the single visual consistency of the same image's different views.

Next, we investigate the effect of erased area on accuracy by changing the value of $\alpha$ and $\beta$.
The upper part of Table \ref{tab:11} summaries the experimental results.
From these results, we obtain the best performance with $\alpha=0.4$ and $\beta=0.3$ for lower-level layers.
Meanwhile, when the erased area is greatly reduced (\emph{i.e},
$\alpha=0.2$ and $\beta=0.1$), 8.5\% mAP and 5.1\% top-1 decreases are observed. This is because the model hard capture the different
semantic information of the object.
Furthermore, given that
higher-level feature is semantic and appearance diversity, we also reduce the size of erased area when DDL is plugged in high-level layers. The results are reported in the lower part of Table \ref{tab:11}, while the findings are still the same as before.


\begin{table}[t]
\footnotesize
  \centering
    \begin{tabular}{{p{2.4cm}p{0.1cm}<{\centering}p{0.1cm}<{\centering}p{0.1cm}<{\centering}p{0.1cm}<{\centering}p{0.4cm}<{\centering}p{0.6cm}<{\centering}}}
    \toprule
    \multicolumn{1}{l}{Method}& \multicolumn{1}{l}{$\mathcal{L}_{ce}$}& \multicolumn{1}{l}{$\mathcal{L}_{st}$} & \multicolumn{1}{l}{DDL} & \multicolumn{1}{l}{$\mathcal{L}_{co}$} & mAP & top-1   \\
    \midrule
     \multirow{2}[2]{*}{Baseline} & \checkmark       &       &       &     & 53.0  & 71.6    \\
                             & \checkmark & \checkmark    &       &       & 55.7 & 73.6    \\
     \midrule
    \multirow{2}[2]{*}{Clustering}  & \checkmark        &       &       &   & 50.3    & 69.3     \\
                      & \checkmark  & \checkmark    &    &      & 54.0  & 73.4   \\
    \midrule
    \midrule
    Baseline+ConsLearn      & \checkmark & \checkmark    & \checkmark     & \checkmark     & 63.2  & 77.4   \\
    Clustering+ConsLearn          & \checkmark & \checkmark     & \checkmark     & \checkmark  & \textbf{64.5}    & \textbf{78.1}    \\
    \bottomrule
    \end{tabular}%
    \vspace{0.1cm}
      \caption{Ablation studies of our proposed method on the unsupervised person re-ID task with DukeMTMC-reID. Unlike the baseline, our clustering generates pseudo labels according to features from a view of the same images and use the same labels for two encoded views to optimize model.}
  \label{tab:ablation}%
\end{table}%

\textbf{Clustering}. We set the model as the baseline when training
the network using only the traditional classification component.
To verify the effectiveness of our clustering, we conduct extensive ablation studies. From experimental results in the upper part of Table \ref{tab:ablation}, it can be easily observed that
the performance of our clustering is worse than that of the baseline model.
The main reason is that
the clustering (w/o ConsLearn) focus on local detailed feature
and thus the learned features from different data flows contains different information, leading to different
clustering results. However, our clustering share the same pseudo label for them.
When consistency learning is also conducted (the lower part of Table \ref{tab:ablation}), our clustering is greatly improved and surpass the baseline model with ConsLearn (Baseline+ConsLearn) by margins (mAP=1.3\% and top-1=0.7\%).
This indicates that our clustering begins to ignores low-level features and focus on semantic information with the help of ConsLearn. In this way, our clustering further enhance the visual consistency by using the same cluster for two encoded features.
Additionally, the softmax-triplet loss also improve the performance of our clustering method, which demonstrates the effectiveness of softmax-triplet loss in our proposed framework.



\begin{figure}[t]
\begin{center}
\includegraphics[width=1.0\columnwidth]{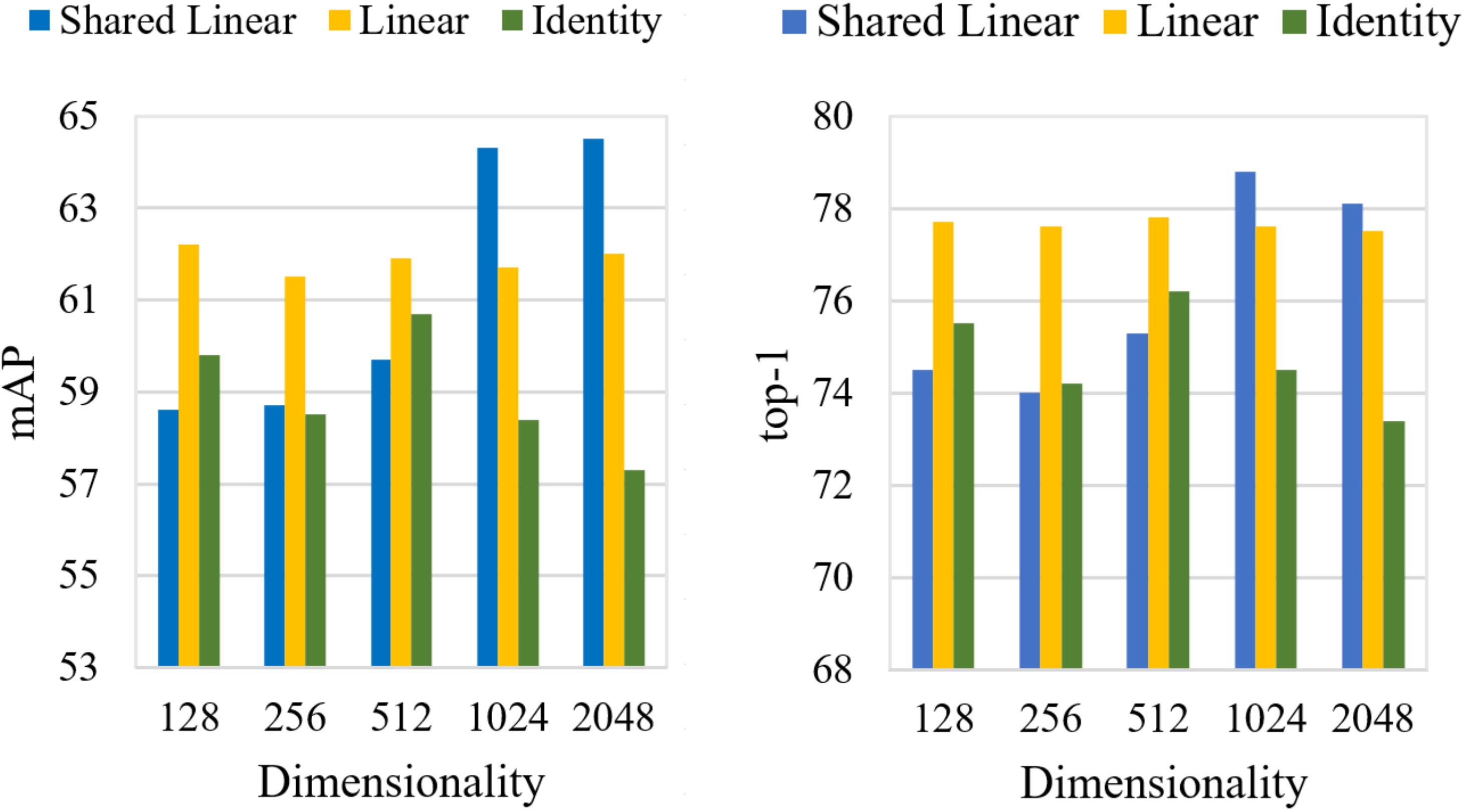}
\end{center}
   \caption{ \small Evaluation results of representations with different projection
heads and various dimensions.}
\label{fig:dim}
\end{figure}

\begin{figure*}[t]
\begin{center}
\includegraphics[width=2.0\columnwidth]{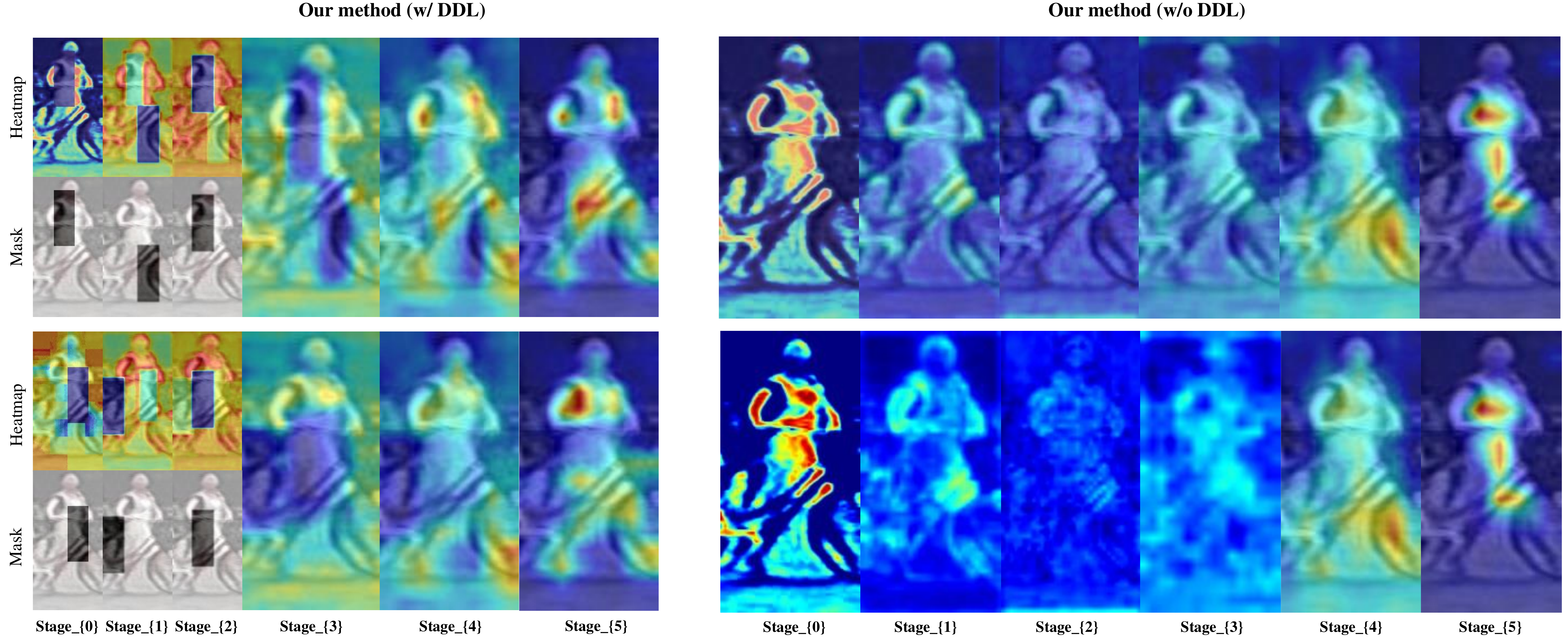}
\end{center}
   \caption{\small The dropmask and heatmap at different layers of ResNet50 \cite{he2016deep} over two encoded features (\emph{i.e.}, $v_{1}$ and $v_{2}$).}
   \vspace{-0.2cm}
\label{fig:heatmap}
\end{figure*}

\textbf{Pretext task}.
As showed in Table \ref{tab:ablation}, the baseline model results in much lower performance than our final method.
A key reason is that the baseline model mainly focuses on the local detailed features to optimize
the target task of re-ID while discarding meaningful semantic feature. As a pretext or base task, consistency
learning is integrated into the baseline model, abbreviated as ``Baseline + ConsLearn". Under this setting, the baseline obtain the obvious improvement (mAP=11.5\% and top-1=6.5\%), which further
demonstrates the importance of adopting LearnCons as pretext task.


\begin{table}[t]
\small
  \centering
    \begin{tabular}{p{1.75cm} p{2.5cm}p{0.64cm}<{\centering}p{0.67cm}<{\centering}p{0.67cm}<{\centering}}
    \toprule
    Method &  Parameter & mAP   & top-1  & top-5   \\
    \midrule
    \multirow{4}[8]{*}{Ours (w/ M)} & $(\theta^{(t)})$  & 64.5  & 78.1 & 87.8
   \\
          & $(\theta^{(t-1)})$ & 63.7  & 78.4  & 87.6  \\
          &  $(\theta^{(t-1)},\theta^{(t)})$ & 63.9  & 77.6 &87.7    \\
          &  $(\theta^{(t-2)},\theta^{(t-1)},\theta^{(t)})$ & 63.9  & 78.2  &87.7  \\
    \midrule
    Ours (w/o M) & N/A    & 59.6  & 74.6  & 85.2    \\
    \bottomrule
    \end{tabular}
    \vspace{0.1cm}
      \caption{\small Evaluation results of the temporally averaging model under different update mechanism.
  M: the temporal averaging model.
  $(\theta^{(t-1)},\theta^{(t)})$: the averaging model is updated as $\theta_{e}^{(t)}= \zeta\theta_{e}^{(t-1)}+\frac{1-\zeta}{2}\theta^{(t-1)}+\frac{1-\zeta}{2}\theta^{(t)}$, and others are similar.}
  \label{tab:momentum}%
\end{table}%

\textbf{Momentum update mechanism}.
The momentum-based averaging model allows us to enforce the temporal consistency over network model at different iterations without
additional model parameter. 
To investigate its effectiveness, we conduct ablation studies by directly use one
network¡¯s current-iteration output as feature distribution, which is abbreviated as ``Ours (w/o M)".
As shown in Table \ref{tab:momentum}, ``Ours (w/ M)" consistently improves the
results over ``Ours (w/o M)", which represents our method without temporal consistency based on averaging model.
The qualitative results demonstrates that the necessary of temporal consistency and the single consistency obtained by model at different stages has different effects on our results.


\textbf{Projection head}. We then study the importance of the linear projections, \emph{i.e.}, $f(\cdot)$ and $g(\cdot)$. Figure \ref{fig:dim} reports evaluation results
using three different architectures for the projection heads: (1) identity
mapping; (2) linear projection; (3) shared linear projection for $f(\cdot)$ and $g(\cdot)$.
We observe that two kinds of linear projection are obviously better than no projection (\emph{i.e.}, identity mapping), especially for the large output dimensions (\emph{e.g.}, 2048 in this paper).
This indicate that, by using two projections, more information can
be formed and maintained in $u_{i}$ and $\tilde{u}_{i}$, $i$=1,2.
Furthermore, the performance of shared linear becomes better as output dimensions increases and outperform normal linear with large dimensions (\emph{e.g.}, 1024 and 2048 in this paper), which demonstrates that shared linear projection can
obtain more information that is useful for two tasks.

\subsection{Discussions}
From the perspective of representation learning, our method integrate disparate visual percepts into persistent entities and underlie visual reasoning in feature space.
As shown in the left of Figure~\ref{fig:heatmap}, the network model (w/ DDL) no longer pays attention to some semantic parts at early stages when they are occluded.
However, at later stages, the predicted target regions of training image are backtraced to the initial information by using the information of the other view regardless of losing the target on a view. Finally, the heatmap extracted
from model almost highlight the entire region of the person.
This indicates that the ability of model to predict the
missing patch of feature maps are enhanced, inching the
two different features toward consistency. On the contrary, the right of Figure \ref{fig:heatmap} shows the model (w/o DDL) only focus on
the most discriminative part of object, not the entire object.
In addition,
unlike the pervious clustering, our clustering
treats the same pseudo labels as ground truth labels for two encoded views to bootstrap the discriminative ability of network model.

Our proposed framework is trained with both
unsupervised clustering and representation learning. In this
way, our method can separate the images into semantic clusters,
where images within the same cluster belong to the semantic classes and images in different clusters are
semantically dissimilar.
Since self-supervised representation learning
is task-agnostic during training, they often
have to focus on the feature information, which is unrelated to the current target task.
Indeed, in our method, the
clustering induces network model to capture feature representation that concentrates on the target task of re-ID. Meanwhile, consistency learning encourages clustering
to obtain semantically meaningful features rather than relying on low-level features.


\section{Conclusion}
In this paper, we present a new approach for unsupervised person re-identification (re-ID) by
combining clustering and consistency learning.
As a target task, clustering learning trains a deep neural network using generated pseudo-labels and concentrates on current task of re-ID. As a pretext task, consistency learning requires semantic understanding and induces clustering to obtain semantically meaningful features rather than depending on low-level features.
Experimental results show that
our method achieves significant improvements over a variety of unsupervised or domain adaptive re-ID tasks.

{\small
\bibliographystyle{ieee_fullname}
\bibliography{egbib}
}

\end{document}